\documentclass[a4paper,10pt,twocolumn]{IEEEtran}%
\usepackage[numbers,sort&compress]{natbib}
\usepackage{varwidth,xcolor}
\usepackage{amsmath}
\usepackage{graphicx}
\usepackage{amssymb}
\usepackage{amsfonts}
\usepackage{algorithm}
\usepackage{times}
\usepackage{fullpage}
\usepackage{placeins}
\usepackage{epstopdf}
\usepackage{float}%
\usepackage{afterpage}
\usepackage{booktabs} 
\usepackage{amsmath}
\usepackage{float}
\usepackage{stfloats}
\usepackage[noend]{algpseudocode}
\makeatletter
\def\algbackskip{\hskip-\ALG@thistlm}
\makeatother

\algnewcommand\RETURN{\algorithmicreturn}
\algnewcommand\PROCEDURE{\item[\algorithmicprocedure]}%
\algnewcommand\algorithmicendprocedure{\textbf{end procedure}}
\algnewcommand\ENDPROCEDURE{\item[\algorithmicendprocedure]}%
\algnewcommand{\algvar}[1]{{\text{\ttfamily\detokenize{#1}}}}
\algnewcommand{\algarg}[1]{{\text{\ttfamily\itshape\detokenize{#1}}}}
\algnewcommand{\algproc}[1]{{\text{\ttfamily\detokenize{#1}}}}
\algnewcommand{\algassign}{\leftarrow}

\algnewcommand\algorithmicswitch{\textbf{switch}}
\algnewcommand\algorithmiccase{\textbf{case}}
\algnewcommand\algorithmicassert{\texttt{assert}}
\algnewcommand\Assert[1]{\State \algorithmicassert(#1)}%
\algdef{SE}[SWITCH]{Switch}{EndSwitch}[1]{\algorithmicswitch\ #1\ \algorithmicdo}{\algorithmicend\ \algorithmicswitch}%
\algdef{SE}[CASE]{Case}{EndCase}[1]{\algorithmiccase\ #1}{\algorithmicend\ \algorithmiccase}%
\algtext*{EndSwitch}%
\algtext*{EndCase}%

\newcommand\dataset[0]{\emph{COFGA} }

\providecommand{\U}[1]{\protect\rule{.1in}{.1in}}
\graphicspath{{./figures/}}
\begin{document}
\pagestyle{plain} 

\title{COFGA: Classification Of Fine-Grained Features In Aerial Images}
\author{Eran Dahan, Tzvi Diskin}

\date{}
\maketitle

\begin{abstract}
Classification between thousands of classes in high-resolution images is one of the heavily studied problems in deep learning over the last decade.
However, the challenge of fine-grained multi-class classification of objects in aerial images, especially in low resource cases, is still challenging and an active area of research in the literature.
Solving this problem can give rise to various applications in the field of scene understanding and classification and re-identification of specific objects from aerial images. 
In this paper, we provide a description of our dataset - \dataset of multi-class annotated objects in aerial images. We examine the results of existing state-of-the-art models and modified deep neural networks. Finally, we explain in detail the first published competition for solving this task.  
\end{abstract}

\section{Introduction}
In recent decades, the amount of footage captured by aerial sensors has been growing exponentially. The abundance of images leads to overflow of information that cannot be processed by human analysts alone, therefore, the absolute majority of the aerial imagery is unlabeled. In those images, one can find images from airborne sensors, satellite sensors, etc... 
The emerging technologies of machine learning and artificial intelligence have begun to address this challenge, and although some progress has been made to solve this task, there are still problems that need to be addressed. One of them is the lack of annotated data to train a deep neural network to automatically classify the different objects as seen from the air, another is the research for a neural network architecture that is suitable to low-resource training, and finally the research to build a representation for the different features that can be jointly learned for different classes (i.e., a mutual representation for the color feature of an object).
There are also factors that affect the ability to solve the task of fine-grained classification from aerial images such as the resolution of the images and the degree of details we wish to resolve.\\ 
Solving this task can give rise to different applications, one of them is the ability to classify objects in order to improve automatic tracking, detailed definition, exploitation, and re-identification.\\
We can conclude our main contribution to three:
\renewcommand{\labelenumii}{\roman{enumi}}
\begin{enumerate}
    \item We collected and annotated the most extensive dataset - \dataset - with high-quality aerial images with fine-grained multi-class annotation. (Section \ref{DATASET})
    \item We published \dataset along with a challenge to develop algorithms for solving this task. (Section \ref{COMP})
    \item We proposed a modified algorithm that was trained on \dataset, and we compare it to known deep networks algorithms that are state-of-the-art for classification. This comparison also serves as a baseline for further research. (Section \ref{RESULTS})
\end{enumerate}

\begin{figure}[t]
    \centering
    \includegraphics[height=3in, width=2.4in]{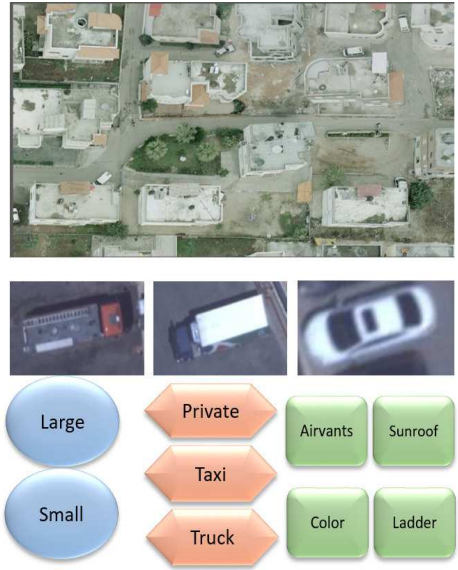}
    \caption{Example image and the object crops taken from it; below is a small part of classes (blue), sub-classes (red) and features (green), note that some object can share features while others cannot by definition.}
    \label{fig:main_idea}
\end{figure}

\section{Related Work}\label{RELATED}

In this section, we describe previous work done in fine-grained classification, further we describe the research done in classifying an object in aerial images.\\
While searching datasets for training deep neural network for the task of fine-grained classification from aerial images, one can find that there are several public databases of labeled aerial imagery. To date, the largest database is xView \cite{XVIEW}; this dataset covers 1,400 square kilometers of satellite images with a resolution of $0.3$ meters GSD. The dataset contains about one million labeled objects divided into different classes. This dataset also has the highest level of specification, which contains a total of 60 categories, including 7 relatively general parent categories ('fixed-wing aircraft', 'passenger vehicle', 'truck', 'railway vehicle', 'engineering vehicle', 'marine vessel', and 'building'), each of which is divided into several sub-categories or specific instances (for example the category ‘truck’ includes the sub-categories: ‘Truck with box trailer’, ‘Truck with flatbed trailer’, ‘Truck with liquid tank’ etc.)\\
Another important database for aerial imagery is DOTA (Dataset for Object Detection in Aerial Images) \cite{DOTA}, which includes almost $3,000$ images with a varying resolution from various aerial sensors and around $190,000$ tagged objects. This dataset refers to 15 different categories of objects and contexts, including planes, bridges, harbors, etc, but does not relate to sub-categories or fine-grained features of these objects.\\
Finally, another dataset we mention is COWC (Cars Overhead With Context) \cite{COWC}; this dataset includes thousands of images from several different aerial sensors (both satellites and aircraft). The dataset was collected from six different regions (e.g., Canada, Germany, USA and New Zealand). However, this dataset was established mainly in favor of the development of object counting algorithms, therefore, its objects are only classified into general categories (e.g., boat, plane, car, etc.)\\
Despite their important advantages, these databases do not contain fine-grained labels in details as needed for fine-grained classification.

\subsection{CNN for image classification}
Image classification is one of the fundamental tasks of computer vision. In the last decade, since the significant improvement in performance on the ImageNet dataset \cite{deng2009imagenet} that was achieved by \cite{krizhevsky2012imagenet},  the object classification task is dominated by convolutional neural networks (CNN) methods. These methods still improve each year \cite{hu2017squeeze}.

\subsection{Aerial object classification}
Despite the extensive work in the field of CNN for image classification, there is no significant work of designing special CNNs for aerial images. Most of the work that was done in this area is applying standard CNN methods to aerial image datasets as in \cite{DOTA} and \cite{radovic2017object}. Interesting related work was done by \cite{dieleman2015rotation} in which a rotation invariant CNN is proposed, and thus may fit well with aerial images.

\subsection{Fine-grained classification}
It is common to distinguish between coarse-grained classification in which an image is assigned to a small set of main classes and fine-grained classification in which each main class is divided to a large number of sub-classes which may be very similar.
Fine-grained image classification is very challenging, mainly because of the small inter-class variance alongside the high intra-class variance due to different pose, scale, rotation, etc. Another great challenge is a big effort that is needed to annotate a large scale and high-quality datasets. A variety of the techniques were developed to address these challenges. \cite{xiao2015application} use an attention mechanism in order to select relevant patches to a certain object for dealing with the small inter-class variations, while \cite{krause2016unreasonable} propose the use of large-scale noisy data in order to significantly decrease the cost of annotating the dataset.

\subsection{Few shot learning}
Machine learning (and especially deep learning) algorithms need many training data in order to perform well. Few shot learning is the case when there are only a few training examples of a desired class (but many examples of other classes).  
In the last years, a lot of work was done in this task. Most of the Methods use different techniques of metric learning \cite{koch2015siamese, vinyals2016matching}, attention mechanism \cite{shyam2017attentive}, data augmentation \cite{hariharan2017low}, or meta learning \cite{vinyals2016matching, bertinetto2018meta}.

\section{fine grained Dataset (\dataset)}\label{DATASET}
\begin{figure*}[t]
\centering
\includegraphics[height=6.4in, width=6.2in]{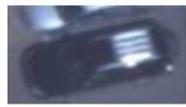}
\caption{Comparing objects type and features over different aerial datasets; A, B, C, D are images from XVIEW, COWC, DOTA and \dataset dataset, respectively. below each image is an object taken from the dataset with its corresponding features}
\label{fig:DATASETS_EX}
\end{figure*}

The dataset we present here is an extensive and high-quality resource that will hopefully enable the development of new and more accurate algorithms for fine grained classification of aerial imagery. Compared to other open source datasets for aerial imagery, it has two notable advantages: First, its resolution is very high (5-15 cm GSD). Second, and most prominent, is that the data is already tagged with fine-grained classifications, referring to delicate and specific characteristics of vehicles, such as air condition vents, the presence of a spare wheel, a sun-roof, and many more.
An illustration of the above comparison can be seen in fig \ref{fig:DATASETS_EX}

\subsection{Empirical Details}
\dataset dataset contains 1,663 images, captured in various land types - urban areas, rural areas and open spaces - on different dates and at different times of the day (all performed in daylight). Images also differ in the size of the covered land area, weather conditions, photographic angle, and lighting conditions (light and shade).
In total, it contains 11,617 tagged vehicles, classified into categories which are divided to four granularity levels:
\begin{itemize}
  \item \textbf{Class}---The category contains only two instances: ‘Large vehicles’ and  ‘Small vehicles’, according to the vehicle's measurements.
  \item \textbf{Sub-class}---'Small' and 'Large' vehicles are divided according to their kind or designation. Small vehicles are divided into a sedan, hatchback, minivan, van, pickup truck, jeep, and public vehicle. Large vehicles are divided into a truck, light truck, cement mixer, a dedicated agricultural vehicle, crane truck, prime mover, tanker, bus, and minibus.
  \item \textbf{Features}---This category deals with the identification of each vehicle’s unique features.
  The features tagged in small vehicles were: sunroof, luggage carrier, open cargo area, enclosed cab, wrecked and spare wheel.
  The features tagged in large vehicles were: open cargo area, AC vents, wrecked, enclosed box, enclosed cab, ladder, flatbed, soft shell box and harnessed to a cart.
  \item \textbf{Object perceived color}---Identification of the vehicle's color: white, grey/silver, blue, red, yellow and other.
\end{itemize}

It should be noted that an object can be assigned with more than one feature, but is assigned to only one sub-class and only one color.

\begin{figure}[t]
    \centering
    \includegraphics[height=5.4in, width=3.1in]{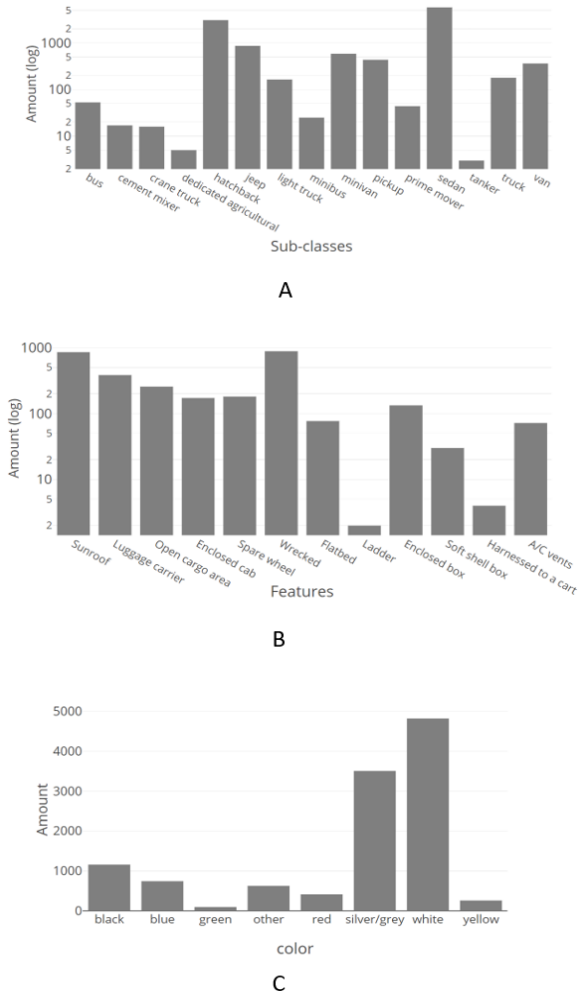}
    \caption{Details of the dataset - distribution of subclasses, features and color shown in A, B, C respectively} 
    \label{fig:DATA_DIST}
\end{figure}

Figure \ref{fig:DATA_DIST} shows the distribution of the sub-classes(A), the features (B) and the colors(C).
figure \ref{fig:DATA_DIST}(A) shows high variance of the sub-classes distribution, \ref{fig:DATA_DIST}(B) that there are two very rare features while the other features have similar number of instances (However the fraction of each of them is less than 10 percent).
figure \ref{fig:DATA_DIST}(C) shows that all colors but white and silver have very low amount of data.

\subsection{Creating the Dataset}
The images were taken with a camera designed for high-resolution vertical and diagonal aerial photography, mounted on an aircraft.

\subsubsection{Phase-1: Initial Labeling}

Two aerial imagery analyst teams first systematically scanned each image and annotated every detectable vehicle with a 4-point bounding box. Each bounding box was labeled as either ‘Small Vehicle’ or ‘Large Vehicle’. Objects that were less than $15\%$ visible (either because they were cut out of image borders or because of intense shading or because clouds obscured them) were omitted. Objects that appeared in more than one image were labeled separately in both image, but such cases were scarce. The bounding boxes were drawn on a local vector layer of each image, thus, their metadata does not entail geographic coordinates. 
The quality of these initial detections and their matching labels was tested at the fine-grained labeling stage by aerial imagery analysis experts. These tests indicated that $4.3\%$ of the detections were false positives, about $4.5\%$ of the labels were incorrect (a small vehicle was labeled as a large vehicle or vice versa) and about $6.2\%$ of the objects were not detected (false negative). All of these cases were disregarded and omitted from the fine-grained analysis phase.

\subsubsection{Phase-2: Fine-grained analysis}

At this point, the team of aerial imagery analysis experts systematically performed fine-grained classification of every annotated object. To ensure the uniformity and accuracy of their work, they first created an analysis-guide manual which contained several representative images for each sub-class, feature, and color. In order to improve the efficiency of this stage, a desktop application was developed to enable a sequential presentation of the labeled bounding boxes and the relevant image section (it also enabled basic analysis properties such as zooming and rotation of the image). For each bounding box, an empty metadata card was displayed, on which the analysts could fill the fine-grained labels. This metadata was recorded to the dataset automatically, but the analysts still had access to the CSV output in the application, which enabled them to verify that they had reviewed all the images and that their labels were accurate. 
After completing the second stage of decoding, a sample of the data was double-checked by an independent aerial imagery expert. This expert performed a systematic check of all the labels of objects belonging to categories that contain relatively few instances ($20$ objects or fewer).

\subsection{Dataset Statistics}
\subsubsection{inter and intra subclass correlation} 
\begin{figure}[h!]
    \centering
    \includegraphics[height=2.3in, width=3.2in]{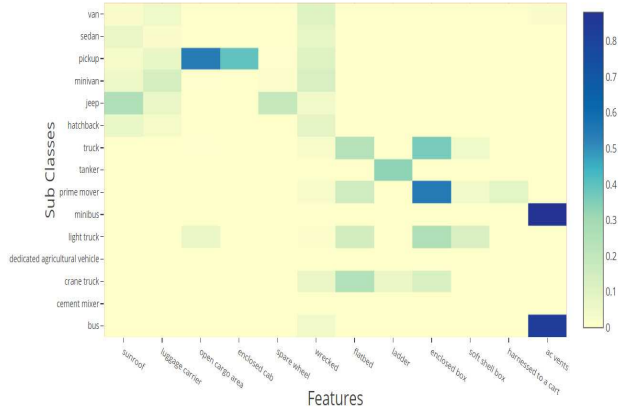}
    \caption{HeatMap of the inter and intra subclass correlation}
    \label{fig:heatmap}
\end{figure}

Each object in the data set is assigned to multi-label vector. These labels are not independent, for example, most of the objects with "spare wheel" are in the sub-class of "jeep".
Figure \ref{fig:heatmap} shows the  inter and intra sub-class correlation

\begin{figure*}[!b]
    \centering
    \includegraphics[height=2.1in, width=6.2in]{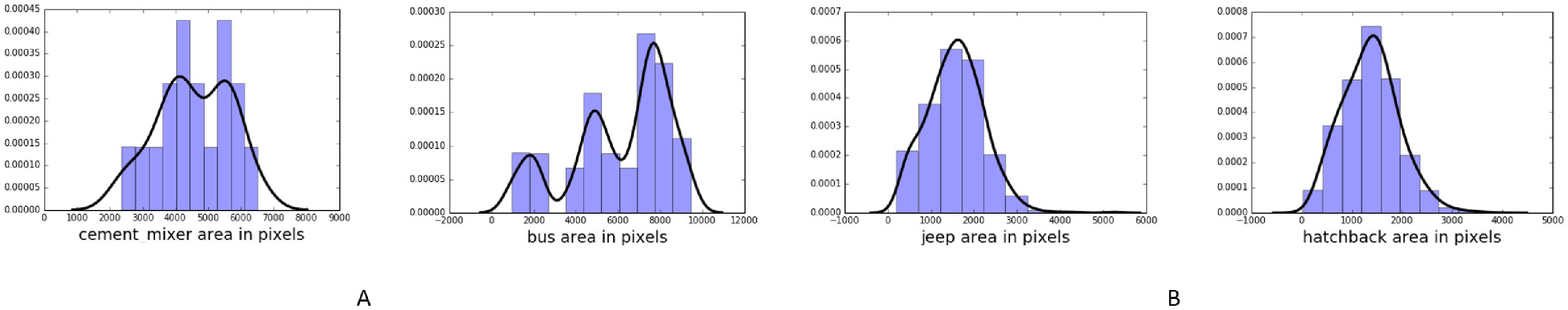}
    \caption{distribution of areas in pixels for different sub-classes from large and small vehicles denoted by A and B respectively}
    \label{fig:areas}
\end{figure*}

The inter sub-class correlation is a measure of the correlation between different sub-classes. From the features point of view, it is the measure of how the same feature is distributed in different sub-classes. Hence, while exploring the heat map above, the inter sub-class correlation can be seen in the value distributed in the columns of the heat map where high value means that the distribution of this feature has a peak for the specific sub-class.
The intra sub-class correlation is a measure of the correlation between different features for a specific sub-class. Hence while exploring the heat map above, the intra sub-class correlation can be seen in the values of the rows of the heat map.
One can see that the most common feature (in being shared through different sub-classes) is the feature of the vehicle being wrecked.
On the other hand, it can be seen that the most correlative sub-class (in having the most significant amount of different features) is a pickup.
Also, the most correlative pair of feature-sub-class is finding a minibus with air condition vents.

\subsubsection{Area of sub classes} 
Another informative statistics that should be considered is the statistics of the sizes for different sub-classes in the dataset.
We compute the distribution of the areas for different sub-classes as can be seen in fig \ref{fig:areas}.
As explained above, we annotated the different sub-classes from 5-15 GSD. While exploring the distribution one can notice that for the large vehicle (e.g., image A) there are usually 2-3 distinct peaks, while for a small vehicle (e.g., image B) the three peaks are coalesced to one.

Also, it is more common finding a different size large vehicles, then finding different size small vehicles, even when those large vehicles are generated from the same subclass (such as finding different size buses compering to finding different size hatchback)
One can use this distribution in order to pre-process the images to better fit and train the classifiers for the task. One can also try to measure the GSD per image and cluster images by GSD in order to better fit the potential area of the specific classifier.
\begin{figure*}[t]
\centering
\includegraphics[height=6.6in, width=6.2in]{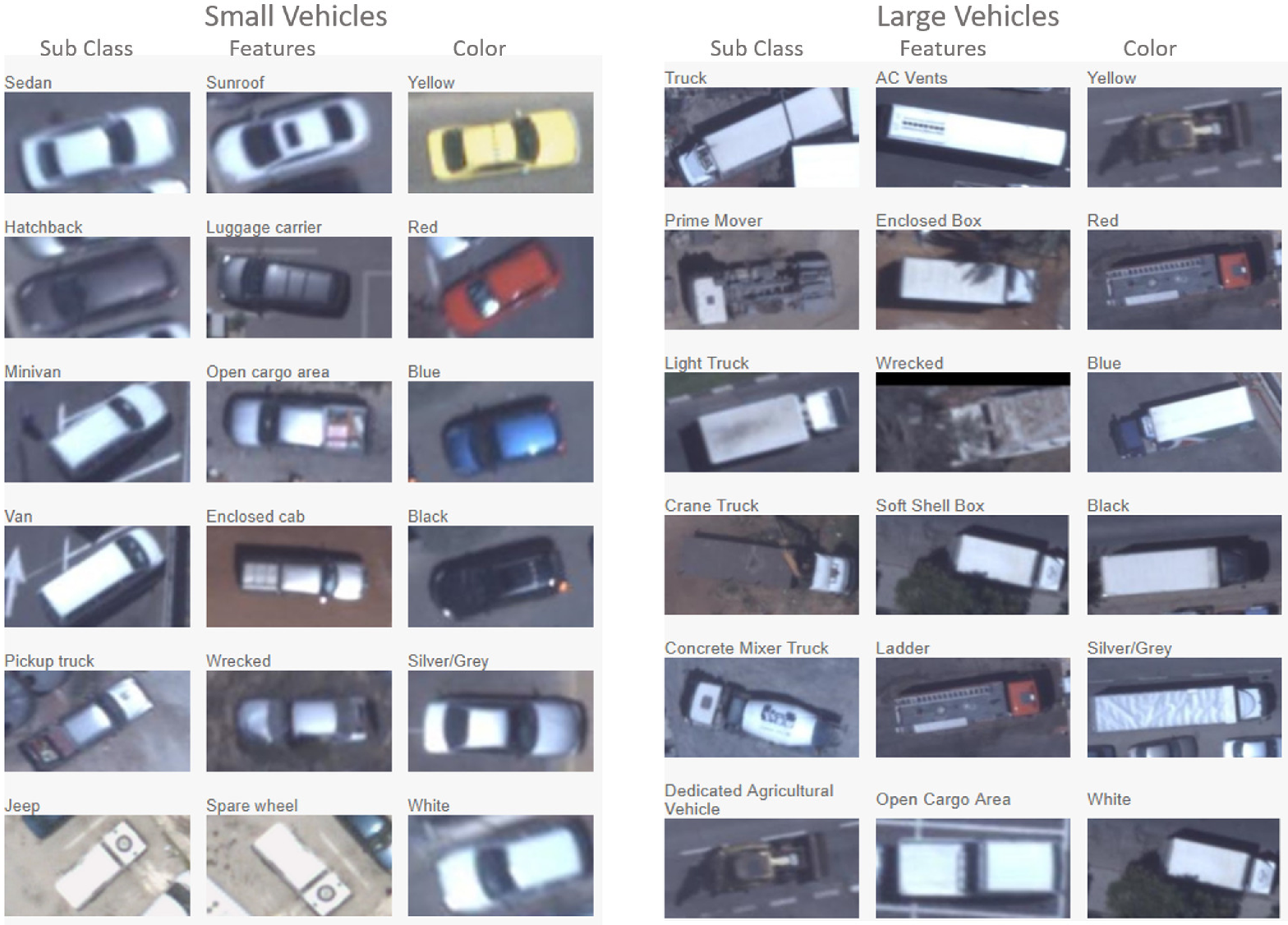}
\caption{Examples from bounding boxes of the competition training set for the different classes subclasses features and colors}
\label{fig:DATASET_EX_LARGE_SMALL}
\end{figure*}
\section{Competition for fine grained classification}\label{COMP}
We published a competition for fine-grained classification of the COFGA dataset on CodaLab. 
The goal of the competition is to develop an algorithm for automatic fine-grained classification of aerial imagery data. 
For the training set, participants received data of $1,697$ \emph{tiff} and \emph{jpeg} images as well as a CSV file of annotated objects. Each object is represented by a tag ID, image ID, and a bounding quadrilateral, which is a set of 4 (x;y) coordinates in the relevant image.
Additionally, each object in the training set includes fine-grained classification labels: class, subclass, features, and color. 
For the evaluation of the algorithm, participants will also receive a test set, consisting of $1,421$ \emph{tiff} and \emph{jpeg} images and a CSV file of non-annotated objects which includes objects in the same form as in the training set (tag ID, image ID, and a bounding quadrilateral), without the classification data.
The competition has two phases - public and private. In the public phase, the submission limit is five per day, and in the private phase, a total of three submissions is allowed.
One of the significant challenges in this competition is that for some sub-classes and features there are only a small number of objects in the training and test set, while other sub-classes and features contain thousand or more tagged objects.
For evaluation of each category, an average precision index is calculated separately. Then, a Quality Index will be calculated as the average of all average precision indices (Mean Average Precision).
The score will be calculated for each category separately according to:
\begin{equation*}
    \text{AP(category)=}\frac{\text{1}}{\text{K}}\sum\limits_{\text{k=1}}^{\text{K}}{\text{Precision(k) rel(k)}}
\end{equation*}

Where K is the total number of objects from the class in the test data.
Precision(k) is the precision calculated over the first k objects and rel(k) equals $1$ if the classification k is correct and $0$ otherwise.\\
To compute the MAP we use:
\begin{equation*}
\text{MAP(system)=}\frac{\text{1}}{{{N}_{c}}}\sum\limits_{\text{category=1}}^{{{N}_{c}}}{\text{AP(category)}}
\end{equation*}
When $N_{c}$ is the number of categories.
Examples from the training set can be seen in figure \ref{fig:DATASET_EX_LARGE_SMALL}.
Every category in the fine-grained classification has the same weight in the total score, therefore, the weight of a small sample size category (e.g. minibus) is equal to a large sample size category (e.g sedan). This index varies between $0$ to $1$ and emphasizes correct classifications with significance to confidence in each classification.

\section{Baseline results for \dataset}\label{RESULTS}
We evaluate state-of-the-art CNN architectures on \dataset. It has some unique properties that we keep in mind for this evaluation:
\begin{enumerate}
    \item \dataset is from an aerial view, and thus it has different symmetry properties than natural images (mainly approximate symmetry to rotations).
    \item The classes in \dataset are highly unbalanced - from few examples for some rare classes to thousands of examples of other classes.
    \item The classes with few training examples are not fit to the current deep learning classification algorithms.
    \item The label of each object consists of multiple attributes which have some correlations between them, for example - a small vehicle cannot have the enclosed box feature.
    \item The objects are annotated by bounding boxes of different size, while the entire image is given.
\end{enumerate}

\subsection{The Proposed Baseline Solution}
The first step in our solution pipeline is to generate images for classification by cropping tiles of the objects according to the labeled bounding boxes. although the objects are annotated by a quadrilateral (8 degrees of freedom), we crop by the appropriate horizontal bounding boxes, using the quadruplet $(x_{min}, y_{min}, x_{max}, y_{max})$

In order to feed the tiles to a standard CNN, we warp the crops to a fixed size of 128 x 128, similarly to \cite{girshick2014rich}. Before the cropping, we dilate the bounding box with 5 pixels of background in each edge.

\begin{figure}[t!]
\centering
\includegraphics[height=2.6in, width=3.1in]{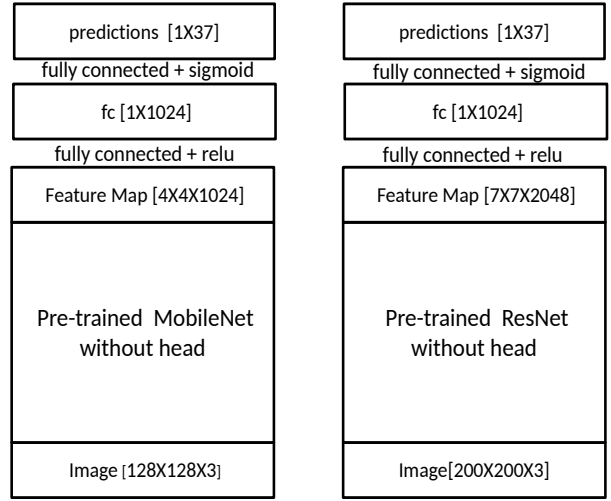}
\caption{Network architectures that were used, based on MobliNet (left) and ResNet50 (right)} 
\label{fig:architectures}
\end{figure}

We generate a $37$ size labels vector for each object. The vector indicates if the object belongs to each of the general classes, sub-classes, colors or features. We treat all these labels as uncorrelated and use binary cross-entropy loss for each of the labels.
We use a pre-trained state-of-the-art CNN architecture and fine-tune it on \dataset dataset. We keep only the convolutional part on the pre-trained network and add a fully connected layer of size $1024$ with ReLU activation, and another fully connected layer of with a sigmoid activation for the final predictions (Figure \ref{fig:architectures}).
We investigate a few different variations to the general pipeline:
\begin{enumerate}
    \item Base Model - we compare the results of using MobileNet \cite{howard2017mobilenets} and ResNet50 \cite{RESNET} architectures.
    \item Weighted Loss - Because of the imbalance between the classes, we assigned a different weight to the loss of each class.
    \item Data Augmentation - We perform a random rotation to the training examples.
    \item Post-Processing - We modify impossible combinations of the classes.
\end{enumerate}

The evaluation metric was chosen to be Mean Average Precision (MAP) of all the general classes, sub-classes, colors and features. As the rare categories are much more challenging on one side and have the same weight as the common categories in the final scores on the other side. The final score is highly dominated by the scores on the rare and the difficult classes.

\subsection{Implantation details}
We use Keras library \cite{chollet2015keras} and its pre-trained models (who were trained on ImageNet \cite{deng2009imagenet}. We use the following parameters: 
\begin{enumerate}
    \item Stochastic gradient descent (SGD) optimizer with a constant learning rate of 0.01 (0.002 for weighted cross entropy loss) and 0.9 momentum.
    \item 100 epochs. 
    \item batch size of 32.
    \item All the layers are fine tuned.
\end{enumerate}

The weighted cross entropy loss is defined as:
\begin{equation*}
    L = \frac{1}{NM}\sum^N_{n=1}[w \cdot y_n \log \hat{y_n} + (1-y_n)  \log (1-\hat{y_n})]
\end{equation*}
where $y_n$ is the labels vector of sample $n$, $\hat{y_n}$ is the vector of predictions of sample $n$, $w$ is the weight vectors, N is the number of samples and M is the length of the labels vector (In our case - $37$)\\
The weight vector was chosen to be:
\begin{eqnarray*}
  w  & = &  (1 - w_0) / w_0 \\
  w_0 & = & \max(\frac{N_m}{N}, 0.1)
\end{eqnarray*}

where $N_m$ is the number of samples in the category $m$. The minimum, value of $0.1$ is set in order to prevent domination of very rare labels.

The post-processing is done by first identifying the main class of a sample by taking the samples where the score of a large vehicle or small vehicle label is more than $0.5$, and then setting all the impossible features' (as set in the training set as -1) scores to zero.  
Because ResNet50 in the Keras framework accepts inputs with a minimum size of 197 x 197, we zero pad the 128 x 128 crop to fit this size.

\begin{table*}[t]
    \centering
    \begin{tabular}{|l|l|l|l|l|}
    \hline
    Base model & Data augmentation & Weighted cross entropy & Post processing & mAp  \\ \hline
    MobileNet  &                   &                        &                 & 0.52 \\ \hline
    MobileNet  & \multicolumn{1}{|c|}{\checkmark}                 &                        &                 & 0.58 \\ \hline
    MobileNet  & \multicolumn{1}{|c|}{\checkmark}               &                        &          \multicolumn{1}{|c|}{\checkmark}       & 0.58 \\ \hline
    MobileNet  &                   & \multicolumn{1}{|c|}{\checkmark}                      &                 & 0.5 \\ \hline
    MobileNet  &       \multicolumn{1}{|c|}{\checkmark}            & \multicolumn{1}{|c|}{\checkmark}                      &                 & 0.6 \\ \hline
    ResNet50  &                    &                        &                 & 0.51 \\ \hline
    ResNet50  &        \multicolumn{1}{|c|}{\checkmark}           &                        &                 & 0.57 \\ \hline
    \end{tabular}
    \caption{evaluation results with different implementations.}
    \label{table 1}
\end{table*}

\begin{table}[h!]
\centering
  \caption{AP for the trained MobileNet network with respect to the classes, sub-classes, features, and number of examples in the training set} 
  \label{tab:Results}
  \begin{tabular}{l c c}
    \toprule
    category & Ap & num of examples \\
    \midrule
     small vehicle &    0.9974 &    11111 \\
     large vehicle & 0.9449 &    506 \\
     minibus &    0.6057 & 25 \\
     hatchback & 0.7637 &    3080 \\
     sedan &    0.9414 &    5783 \\
     bus &    0.9597 &    53 \\
     minivan &    0.3478 &    586 \\
     truck &    0.6002  &    179 \\
     van &    0.7551 &    362 \\
     jeep &    0.1379  &    865 \\
     cement mixer &    0.2824  &    17 \\
     dedicated agricultural vehicle &    1.0000  &    5 \\
     tanker &    0.0499  &    3 \\
     crane truck &    0.8333  &    16 \\
     pickup &    0.5329  & 435 \\
     light truck &    0.1599  &    164 \\
     prime mover &    0.6706  &    44 \\
     red &    0.9283  &    414 \\
     black &    0.8696  &    1158 \\
     blue &    0.6491  &    742 \\
     silver/grey &    0.8026   &    3505 \\
     white &    0.9321  &    4817 \\
     other &    0.1355  &    626 \\
     yellow &    0.9445 &    258 \\
     green &    0.3321  &    97  \\
     sunroof &    0.8904 &    853 \\
     luggage carrier &  0.7984  &    383\\
     open cargo area &     0.5905  &    256 \\
     enclosed cab &    0.5205  &    172 \\
     spare wheel &    0.3439  &    181 \\
     wrecked &    0.4210  &    881 \\
     flatbed &    0.5329  &    77 \\
     ladder &    0.0311  &    2 \\
     enclosed box &    0.6546  &    133 \\
     soft shell box &    0.3624   &    30 \\
     harnessed to a cart &    0.0192  &    4\\
     ac vents &  0.9439 &    72 \\
\end{tabular}

\end{table}

\subsection{Results}
We evaluate different combinations of the methods that were described in the previous subsection.
We evaluate MobileNet with and without data augmentation, with regular cross-entropy loss and with weighted cross-entropy loss. We checked basic post-processing. We also evaluate ResNet50, but because we saw that the results are very similar to those of MobileNet, we did not check all the combinations. Full results are shown in table \ref{table 1}.

\begin{figure*}[t!]
    \centering
    \includegraphics[height=5.0in, width=6.2in]{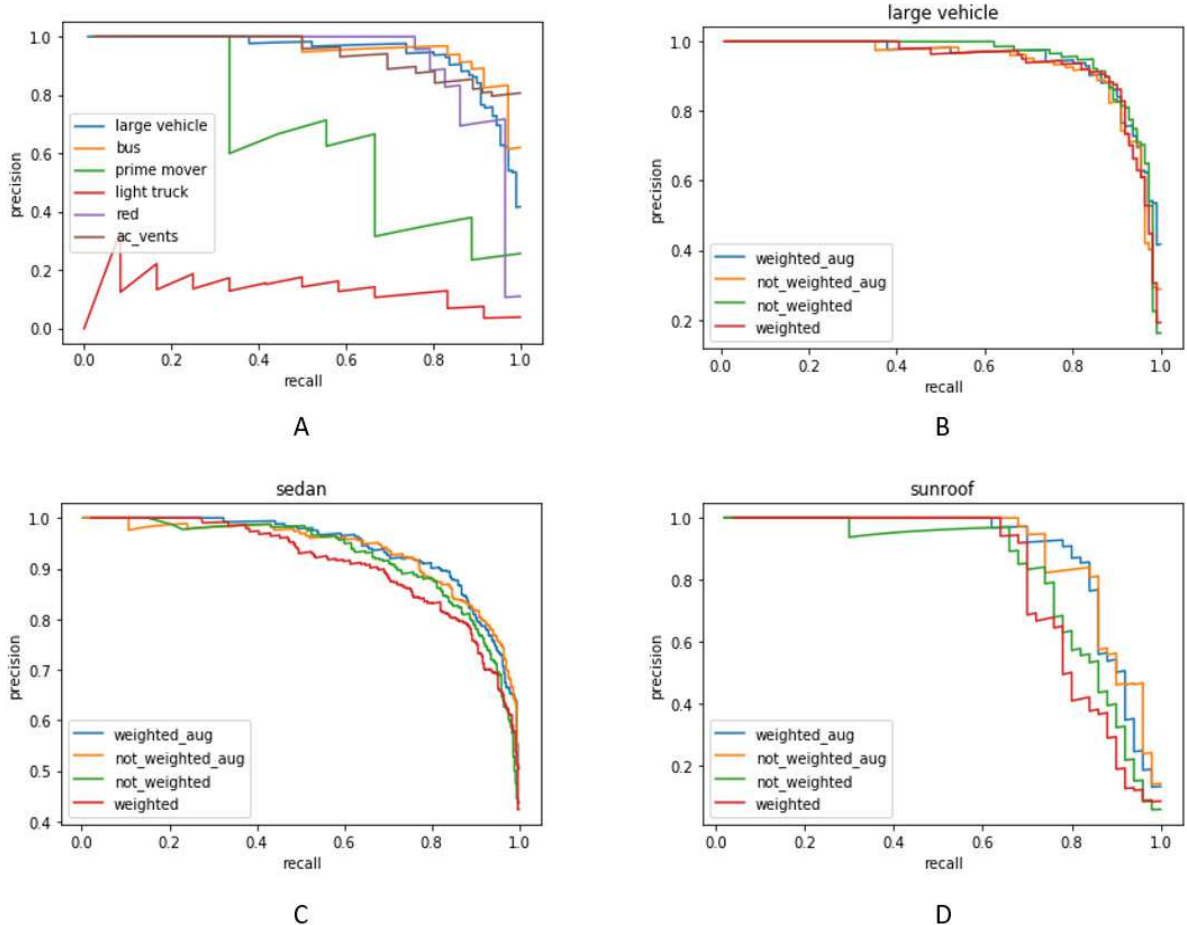}
    \caption{Test evaluation; A - Mobile net precision vs. recall; B, C, D - precision vs. recall for different methods on a class, subclass and features respectively}
    \label{fig:areas_f}
\end{figure*}

Table \ref{tab:Results} shows the AP score on the different categories using MoblieNet that was trained with
data augmentation and weighted cross-entropy loss. Some categories achieve almost perfect results (e.g. "sedan"), some categories achieve fair results (e.g., "truck") while some categories achieve poor results (e.g., "ladder"). 
Figure \ref{fig:areas_f} (A) shows the recall-precision curve of MoblieNet that was trained with data augmentation and weighted cross-entropy loss, on different categories. Some categories ("large vehicle", "bus", "red" and "Ac vents") get very good results. "Tanker" gets high precision on low recall, which means that the first predictions for it are correct and "light truck" gets low precision at low recall, which means that even the first predictions are likely wrong.
Figure \ref{fig:areas_f} (B-D) shows The recall-precision curve on "large vehicle", "sedan" and "sunroof" respectively, with different models. It can be seen that the models that were trained with data augmentation ("weighted aug" and "not weighted aug") achieve better results in all classes, while the influence of the type of the loss is not clear. 

The main conclusions from our evaluation are:
\begin{enumerate}
    \item Data augmentation improves the results significantly.
    \item Base network architecture - MobileNet and ResNet50 achieve very close results. However, MobileNet is much quicker to train.
    \item The influence of weighted cross-entropy is not significant (and not consistent).
    \item Basic post-processing does not help. 
\end{enumerate}

\section{Conclusion and future work}\label{CONCLUSION}
In this paper, we investigated the challenges in a fine grained object classification from aerial images. We explored the state-of-the-art classification algorithms and showed their insufficient results of those algorithms based on our annotated dataset - \dataset.
We offered \dataset dataset for public use and further explained the objects and different classes in it.
We hope that this dataset along with the competition for fine grained classification will give rise to developing algorithms in this area.
At this point, we applied fine-grained classifications only to vehicles. However, we hope that in the near future this trend will develop and apply to other situations, contexts, and objects appearing in aerial photographs, other than vehicles. We believe that the high-resolution images and the accurate and fine-grained classifications turn our proposed dataset to fertile ground for the development of technologies for automatic fine-grained analysis and labeling of aerial imagery. 
Such labeling can help further on, with the challenge of semantic retrieval of data from aerial imagery. Today, retrieval questions such as: “looking for all cases where a red minivan with a sunroof and a spare wheel, appeared in a certain area within a specific time frame” are hard to do, mainly because aerial imagery are, at large, unlabeled. Such retrieval capabilities will open up many new uses of data from aerial imagery.

\FloatBarrier

\bibliographystyle{IEEEbib}
\bibliography{MAIN.bib}


\end{document}